\documentclass[conference]{IEEEtran}
\IEEEoverridecommandlockouts
\usepackage{cite}
\usepackage{amsmath,amssymb,amsfonts}
\usepackage{graphicx}
\usepackage{multirow}
\usepackage{textcomp}
\usepackage{hyperref}
\usepackage{booktabs}
\usepackage{xcolor}

\usepackage{algorithm}
\usepackage{algorithmic}
 
\makeatletter
\newcommand{\removelatexerror}{\let\@latex@error\@gobble}
\makeatother

\def\BibTeX{{\rm B\kern-.05em{\sc i\kern-.025em b}\kern-.08em
    T\kern-.1667em\lower.7ex\hbox{E}\kern-.125emX}}
\begin{document}

\title{A Multi-Granularity Supervised Contrastive Framework for Remaining Useful Life Prediction of Aero-engines}

\author{\IEEEauthorblockN{1\textsuperscript{st} Zixuan He}
\IEEEauthorblockA{\textit{College of Cont. Sci. and Eng.} \\
\textit{Zhejiang University}\\
Hangzhou, China \\
22160009@zju.edu.cn}
\and
\IEEEauthorblockN{2\textsuperscript{rd} Ziqian Kong}
\IEEEauthorblockA{\textit{China-Austria Belt $\&$ Road Joint Lab on AI $\&$ Adv. Manuf.} \\
\textit{Hangzhou Dianzi University}\\
Hangzhou, China \\
kongziqian94@163.com}
\and
\IEEEauthorblockN{3\textsuperscript{th} Zhengyu Chen}
\IEEEauthorblockA{\textit{College of Cont. Sci. and Eng.} \\
\textit{Zhejiang University}\\
Hangzhou, China \\
chenzy1999@zju.edu.cn}
\and
\IEEEauthorblockN{4\textsuperscript{th} Yuling Zhan}
\IEEEauthorblockA{\textit{College of Cont. Sci. and Eng.} \\
\textit{Zhejiang University}\\
Hangzhou, China \\
sarajlbb@zju.edu.cn}
\and
\IEEEauthorblockN{5\textsuperscript{th} Zijun Que}
\IEEEauthorblockA{\textit{College of Cont. Sci. and Eng.} \\
\textit{Zhejiang University}\\
Hangzhou, China \\
quezijun@163.com}
\and
\IEEEauthorblockN{6\textsuperscript{th} Zhengguo Xu*}
\IEEEauthorblockA{\textit{College of Cont. Sci. and Eng.} \\
\textit{Zhejiang University}\\
Hangzhou, China \\
xzg@zju.edu.cn}
}

\maketitle

\begin{abstract}
Accurate remaining useful life (RUL) predictions are critical to the safe operation of aero-engines. Currently, the RUL prediction task is mainly a regression paradigm with only mean square error as the loss function and lacks research on feature space structure, the latter of which has shown excellent performance in a large number of studies. This paper develops a multi-granularity supervised contrastive (MGSC) framework from plain intuition that samples with the same RUL label should be aligned in the feature space, and address the problems of too large minibatch size and unbalanced samples in the implementation. The RUL prediction with MGSC is implemented on using the proposed multi-phase training strategy. This paper also demonstrates a simple and scalable basic network structure and validates the proposed MGSC strategy on the CMPASS dataset using a convolutional long short-term memory network as a baseline, which effectively improves the accuracy of RUL prediction.
\end{abstract}


\section{Introduction}
Aero-engines gradually experience fatigue and degradation over long periods, leading to a decline in performance and reliability. Therefore, accurate remaining useful life (RUL) predictions are important for timely maintenance and safety improvement.

The current RUL prediction approaches can be categorized into two main groups: physical model-based approaches \cite{8709986} and data-driven approaches \cite{9801529}. Physical model-based approaches use the understanding and modeling of the physical behavior of a system to predict the RUL of a component or device. However, as the complexity of devices increases, it becomes increasingly difficult to build physical models. With the development of technologies such as sensors and the Internet of Things, it has become easier to acquire and save data, which has led to the growing popularity of data-driven RUL prediction approaches.

\begin{figure}[!t]
  \centering
  \includegraphics[width=1\linewidth]{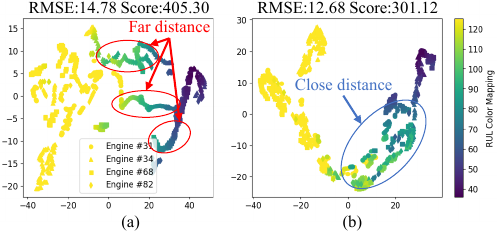}
  \caption{Encoder feature space of four test engines from FD001 visualized by t-SNE. (a) is the feature space for regression training only, and the features with the same RUL label between different engines are far away from each other. (b) is the feature space trained by multi-granularity supervised comparisons with close feature distances between different engines with the same RUL label.}
  \label{fig:{Schematic}}
\end{figure}

Data-driven approaches extract valid features directly from data and use them for RUL prediction without physical knowledge. These can be further categorized into shallow machine learning and deep learning. The former requires manual feature selection and feeding them into predictors such as support vector regressior \cite{XUE202095} to infer RUL. On the contrary, the deep learning approaches adopt an end-to-end training strategy, which simultaneously realizes feature extraction and RUL reasoning by optimizing the loss function through stochastic gradient descent, and has great versatility and flexibility. The best-known deep learning approaches are convolutional neural networks (CNN) \cite{9849459} and long short-term memory neural networks (LSTM) \cite{8967059}. 

However, most of the current deep learning-based RUL prediction approaches use only the mean square error (MSE) as the loss function, a metric that focuses only on the magnitude of the error without considering the underlying structural features, as shown in Fig \ref{fig:{Schematic}}a. Contrastive learning is an unsupervised learning method that uses contrastive loss (e.g., infoNCE \cite{oord2018representation} and triplet loss \cite{weinberger2009distance}) to pull similar embeddings together and push dissimilar embeddings apart in feature space. This helps to improve the performance of downstream tasks \cite{chen2020simple}. For labeled samples, supervised contrastive learning can be used. It extends a single positive sample in minibatch into multiple positive samples by the idea of class label-based aggregation, which results in tighter embedding of different classes \cite{khosla2020supervised}. 

However, for RUL prediction tasks with dense labels, a huge batch size is required to balance positive and negative samples in supervised contrastive learning using RUL labels directly, which is usually limited by hardware. To solve this problem, this paper proposes a multi-granularity supervised contrastive (MGSC) framework that incorporates two contrastive strategies, coarse-grained and fine-grained, to balance the samples while avoiding large batch sizes through a large classification scale label of health status (HS). The intuition of the framework is that the features used as input to the RUL regression layer should be aligned by RUL labels. Based on this, a simple scalable network structure was designed for validation in \ref{struc}. Moreover, to integrate the MGSC and regression tasks, a multi-phase training strategy is proposed. The proposed framework exceeds the accuracy of the baselines on the CMAPSS dataset, as shown in Fig \ref{fig:{Schematic}}b.

The main contributions of this paper are as follows:
\begin{enumerate}
    \item An MGSC strategy is proposed to regularize the feature space and also to solve the problems of large minibatch size and imbalance of positive and negative samples.
    \item A multi-phase training strategy is proposed to progressively achieve the alignment of embeddings of the same RUL samples in the feature space and the regression prediction of RUL.
    \item The proposed MGSC framework effectively improves the prediction accuracy in the baseline.
\end{enumerate}

The paper is organized as follows. Section II describes the proposed MGSC framework, including contrast strategy, training strategy, and network structure. Section III introduces the data preprocessing and model construction. Section IV verifies the validity of the MGSC framework through experimental comparisons. The conclusion is presented in Section V.

\section{Proposed MGSC Framework}
In this paper, the MGSC framework is developed to simultaneously address the conflict between batch size and sample balance as well as the need for feature space structuring in the aero-engine RUL prediction task. It consists of three main parts. The first part is the MGSC strategy to help the encoder network obtain more discriminative and generalizable feature representations with different hierarchical label information. The second part is the multi-phase training strategy to progressively regularize the target feature space, where the embeddings of the same RUL label are aligned. The other part is the base network structure with encoder, projector, and regression layers.

\subsection{Multi-granularity Supervised Contrastive Strategy}
The proposed MGSC strategy incorporates two supervised contrastive forms. For the coarse-grained contrast, the HS serves as the class label, capturing the overall condition of the system. Instead, the fine-grained contrast employs the RUL labels within subsets of each HS. This multi-granularity strategy allows the model to capture roughly high-level semantic information and low-level local variations. Notably, the RULs are real labels reflecting the remaining lifespan of the system, while the HS are pseudo-labels obtained through a segmentation mapping of the RULs, more details are in the \ref{Data preprocessing}. 

In the coarse-grained contrast, the HS are used as class labels. This aims to use larger classification scales for supervised contrastive learning, which enables preliminary regularization of embeddings at smaller minibatch sizes compared to directly using RUL labels (units, i.e., the smallest classification scale). For instance, consider $N$ objects, where each object has $m$ samples. In a given object, contains $d$ HS classes, with each HS class containing $m/d$ samples. On the other hand, there are $N$ RUL classes with each RUL class consisting of only one sample. The number of positive samples for both are respectively:
\begin{align}
    N_{\text{HS}}&=\underbrace{m/d-1}_{\text{Own}} + \underbrace{(N-1)*m/d}_{\text{Others}} \\
    N_{\text{RUL}}&=\underbrace{0}_{\text{Own}} + \underbrace{N-1}_{\text{Others}}
\end{align}
where usually $d\ll m$. Using HS as a class label produces more positive samples that are more likely to be sampled into a minibatch. Therefore, even a small batch size ensures that there are enough positive samples to participate in supervised contrastive training, avoiding the imbalance of positive-negative samples.

The coarse-grained contrastive strategy, as shown in Fig. \ref{fig:{ph}}a, drives to keep the embeddings of the same class of HSs close together, while the embeddings of different classes of HSs away from each other in the feature space. Furthermore, the strategy incorporates samples from both inter- and intra-wise engine views, resulting in a more diverse set of samples in the minibatch. This diversity aids in learning more discriminative and generalized representations.

\begin{figure}[!t]
  \centering
  \includegraphics[width=1\linewidth]{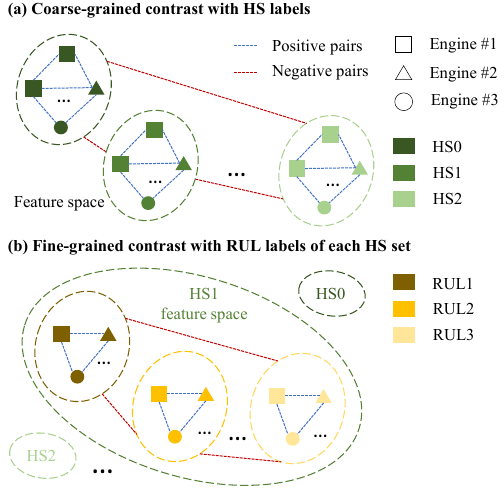}
  \caption{Multi-granularity supervised contrastive learning strategy. (a) The coarse-grained strategy uses HS class labels for contrastive learning and contains both inter- and intra-wise engine views. (b) The fine-grained strategy uses RUL labels in the same HS class for contrastive learning and includes an inter-wise engine view.}
  \label{fig:{ph}}
\end{figure}

However, relying solely on coarse-grained contrasts based on HS labels results in disordered clusters. To ensure that the samples within each cluster can be aligned according to the RUL, as shown in Fig. \ref{fig:{ph}}b, a fine-grained supervised contrastive strategy using RUL labels is designed. In fine-grained contrast, each HS class is processed independently and its sample space is the sample set corresponding to that HS. Since the samples of a single HS class are limited, the minibatch size is completely acceptable and efficient for training. In a specific HS feature space, embeddings with the same RUL label are close to each other, while those with different labels are far from each other. This contrast includes an inter-wise engine view. Moreover, due to the preliminary regularization of coarse-grained, each sample in that specific HS set can be regarded as a hard-negative sample of its anchors, which can improve the robustness of the model.

Under the MGSC strategy, the embeddings of the same HS labels are close together, and the embeddings of the same RUL labels are aligned within that HS cluster. This intuitively implements the feature space regularization.

\subsection{Multi-phase Training Strategy}
Each granularity of MGSC has a different design purpose, as well as the respective minibatch size, so the parallel training paradigm is not used. Instead, this paper proposes a multi-phase training strategy to incorporate supervised contrastive learning with different multi-granularity and RUL regression prediction tasks in a sequential manner.

In a training epoch, coarse-grained supervised contrastive learning is performed first, followed by fine-grained supervised contrastive learning, and finally regression prediction. For coarse-grained contrast, it can roughly regularize embeddings at a large classification scale. Its loss function is:

\begin{align}
    \mathcal{L}_{\text{HS}}=\sum_{i\in I}\frac{-1}{|P(i)|}\sum_{p\in P(i)}\log\frac{\exp(z_i\cdot z_p/\tau)}{\sum_{a\in A(i)}\exp(z_i\cdot z_a/\tau)}
\end{align}
where $z_i=Proj(Enc(x_i))$. And $I$ is the sample space and $i$ is the index of a sample. Here, $P(i)\equiv \left\{p\in A(i): \widetilde{y}_p=\widetilde{y}_i\right\}$ is the set of indices of all positives in the minibatch distinct from $i$, and $|P (i)|$ is its cardinality. $A(i)\equiv I \backslash \left\{i\right\}$ is the sample space without $i$-th sample. 

For fine-grained contrast, it is able to achieve ordered alignment of embeddings within HS clusters at smaller scales. For the particular class $\text{HS}_k$, its loss function is:
\begin{align}
    \mathcal{L}^{\text{HS}_k}_{\text{RUL}}=\sum_{j\in J_k}\frac{-1}{|H_k(j)|}\sum_{h\in H_k(j)}\log\frac{\exp(z_j\cdot z_h/\tau)}{\sum_{b\in B_k(j)}\exp(z_j\cdot z_b/\tau)}
\end{align}
where $J_k$ is the sample space of $\text{HS}_k$ and $j$ is the index of a sample. Here, $H_k(j)\equiv \left\{h\in B_k(j): \widetilde{y}^k_h=\widetilde{y}^k_j\right\}$ is the set of indices of all positives in the minibatch of $\text{HS}_k$ distinct from $j$, and $|H_k(i)|$ is its cardinality. $B_k(j)\equiv J_k \backslash \left\{j\right\}$ is the sample space of $\text{HS}_k$ without $j$-th sample. 

For regression prediction, the loss function is the MSE:
\begin{align}
   \mathcal{L}_{\text{reg}} = \frac{1}{n}\sum^n_{i=1}(l_i-\widetilde{l}_i)^2
\end{align}
where $n$ is the batch size, $l_i$ is the actual RUL, and $\widetilde{l}_i$ is the predicted RUL.

\subsection{Network Structure}\label{struc}
For the proposed MGSC framework, its basic network structure is shown in Fig \ref{fig:{model}}. It includes an encoder to extract the feature embeddings of the samples, a projector to obtain compact embedding representations and a regression layer to predict the RUL. The projector is used only when training, where its projection in the low-dimensional space is used for MGSC training, while the encoder embedding is used directly in regression layers when testing. The overall algorithm is illustrated in Algorithm \ref{alg:1}.

\begin{figure}[!t]
  \centering
  \includegraphics[width=1\linewidth]{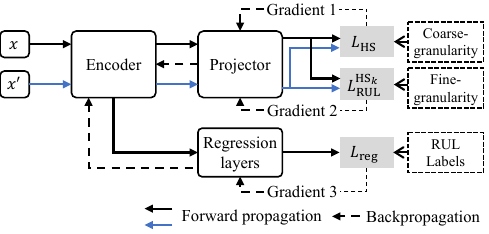}
  \caption{Model structure and process. The two samples were input to the encoder, and the outputs were mapped by the projector into the low-dimensional space. In this projection space, coarse- and fine-grained supervised contrast learning was performed sequentially. Then, regression training was applied using the structured encoder and RUL labels.}
  \label{fig:{model}}
\end{figure}

\begin{figure}[htbp]
    \removelatexerror
    \begin{algorithm}[H]
        \caption{The MGSC framework}
        \begin{algorithmic}[1]
        \label{alg:1}
            \REQUIRE A training contrative sample pair $(x,x')$, corresponds to the RUL label $l$ of $x$, a testing sample $x_t$.         
            \ENSURE RUL prediction $\widetilde{l}$.  
            
            \textbf{Training stage:}
            \STATE $(z,z')=Proj(Enc(x,x'))$.
            \STATE Optimization $\mathcal{L}_{\text{HS}}(z,z')$.
            \STATE Optimization $\mathcal{L}^{\text{HS}_k}_{\text{RUL}}(z,z')$.
            \STATE $r=Enc(x)$.
            \STATE $\widetilde{l}=Reg(r)$.
            \STATE Optimization $\mathcal{L}_{\text{reg}}(l,\widetilde{l})$.

            \textbf{Testing stage:}
            \STATE $r_t=Enc(x_t)$.
            \STATE $\widetilde{l}=Reg(r_t)$.
            \RETURN $\widetilde{l}$ 
        \end{algorithmic}
    \end{algorithm}
\end{figure}

\section{Data Preprocessing and Model Construction}
\subsection{CMAPSS Dataset}
To assess the predictive performance of the proposed model, this research conducted experiments utilizing the CMAPSS simulation dataset provided by NASA, including four subsets. Each sub-dataset consists of a training set and a test set, as well as a set of RUL labels corresponding to the test set. The dataset contains 21 sensor sequence data and 3 operating parameters, which simulate the full life cycle failure degradation of the engine under different operating conditions and failure modes, as detailed in TABLE \ref{tab:cmapss}.

\begin{table}[!t]
\centering
\caption{\label{tab:cmapss}CMAPSS  Dataset}
\begin{tabular}{ccccc}
\toprule
Subsets & Trainset & Testset & Fault model & Woring condition \\ \midrule
FD001   & 100              & 100             & 1           & 1                \\
FD002   & 260              & 259             & 1           & 6                \\
FD003   & 100              & 100             & 2           & 1                \\
FD004   & 260              & 248             & 2           & 6                \\ \bottomrule
\end{tabular}
\end{table}

\subsection{Normalization and Sliding Time Windows}
\subsubsection{Normalization}
Since there are large differences in order of magnitude between different sensor data, mapping them to similar scales through normalization ensures that they have the same importance during model training to avoid bias. The raw data is first normalized: 
\begin{align}
   x_\text{norm}^{i,j}=\frac{2(x^{i,j}-x^j_\text{min})}{x_\text{max}^j-x_\text{min}^j}-1 \quad \forall i,j
\end{align}
where $x^{i,j}$ represents the $i$-th raw data of the $j$-th sensor, $x_\text{norm}^{i,j}$ is the normalized value, and $x_\text{max}^j$ and 
$x^j_\text{min}$ are the maximum and minimum values of the $j$-th sensor, respectively.
\subsubsection{Sliding Time Windows}
The sliding time window technique is applied to obtain sequence data for the sequence regression task. In this paper, we set the window width W=30 and moving step s=1. There is an overlap between the windows and each window corresponds to a RUL label. By this technique, sensor information over a period of time can be obtained, which helps in feature extraction in the temporal dimension.

\subsection{Supervised Label Setting}
\subsubsection{RUL Regression Labels}
In this paper, the normal operation phase of an aero-engine is defined as the period in which the remaining useful life (RUL) exceeds 125 units. During this phase, the engine experiences minimal performance degradation, and the RUL label is maintained at a constant value of 125. Subsequently, as the engine enters the degradation phase, the RUL label gradually decreases linearly from 125 to 0 with the operational time, as shown in Fig \ref{fig:{HS}}.

\begin{figure}[!t]
  \centering
  \includegraphics[width=0.65\linewidth]{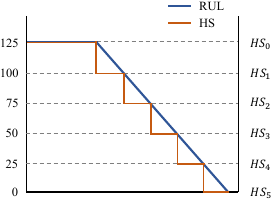}
  \caption{Supervised contrastive labels of HS and RUL.}
  \label{fig:{HS}}
\end{figure}

\subsubsection{Coarse-grained HS Labels}\label{Data preprocessing}
For the HS labels, we have defined different HSs based on the RUL values, as shown in Fig \ref{fig:{HS}}. HS0 is assigned when the RUL is equal to 125. HS1 is assigned when the RUL is between 100 and 124. HS2 is assigned in the range of 75 to 99. HS3 is assigned in the range of 50 to 74. HS4 is assigned in the range of 25 to 49. Finally, HS5 is assigned when the RUL is between 0 and 24.

\subsubsection{Fine-grained RUL Labels}
In fine-grained supervised contrastive learning, its sample space is the subclasses divided in the overall sample space using HS labels. Its labels are the RUL labels corresponding to the samples within each HS class.

\subsection{Model Construction}
In this paper, the CNN-LSTM model has been used as a baseline to compare its RUL prediction performance with or without MGSC strategy training. A concrete implementation of the model construction is shown in Fig. \ref{fig:{model_1}}. The encoder contains three 1D CNN layers for channel feature extraction and utilizes an LSTM module for temporal dimension feature aggregation. The regression layers contain three linear layers, where the output of the last linear layer is mapped to between 0 and 1 utilizing a sigmoid function for normalized RUL prediction. The projector uses two linear layers to project the embedding into a low-dimensional space for MGSC training, where the shallow linear layer uses a Relu activation function for nonlinear transformation \cite{chen2020simple}.

\begin{figure}[!t]
  \centering
  \includegraphics[width=1\linewidth]{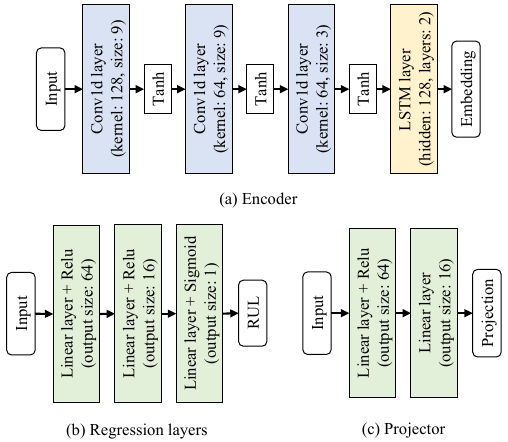}
  \caption{Model Construction. (a) The encoder with three 1D convolutional layers for channel feature extraction and an LSTM module for temporal feature aggregation. (b) The regression layers with three linear layers and the output of the last layer is mapped to between 0 and 1 by a sigmoid function for the normalized RUL prediction. (c) The projector with two linear layers and a Relu function, where the shallow linear layer realizes a nonlinear transformation using the Relu function.}
  \label{fig:{model_1}}
\end{figure}

\section{Experiments}
\subsection{Evaluation metrics}
To evaluate the predictive performance of the model, two commonly adopted metrics, namely root mean square error (RMSE) and scoring function (Score):
\begin{align}
   \text{RMSE}&=\sqrt{\frac{1}{n}\sum^n_{i=n}\Delta_i^2} \\
   \text{Score}&=\left\{
    \begin{aligned}
        &\sum^n_{i=1}\left(\exp\left(-\frac{\Delta_i}{13}\right)-1\right), && \text{if } \Delta_i \leq 0, \\
        &\sum^n_{i=1}\left(\exp\left(\frac{\Delta_i}{10}\right)-1\right), && \text{if } \Delta_i > 0.
    \end{aligned}
    \right.
\end{align}
where $\Delta_i=\widetilde{l}_i-l_i$  is the error between the predicted result and the ground truth.

\subsection{Results}
In this paper, the CNN-LSTM model, CNN model and LSTM model have been used as three baselines to compare the RUL prediction performance with or without MGSC strategy training. Notably, the CNN model is a direct replacement of the LSTM in CNN-LSTM with a linear layer and the LSTM model use multi-layer LSTM as encoder. Table \ref{tab:rul_1} shows that for the CNN-LSTM model and the LSTM model, there are significant improvements in the RUL prediction performance for all four data subsets after using MGSC. For the CNN model, all of them improve except FD004 whose Score metric becomes higher. One of the reasons for the reduced performance may be because FD004 has the most complex operating conditions and failure modes, thus producing a negative contrastive effect. But in general, the proposed MGSC effectively improves the performance of the three baselines.

\begin{table}[!t]
\centering
\caption{\label{tab:rul_1}RUL prediction metrics of MGSC}
\begin{tabular}{cccccc}
\toprule
Approach                       & Metrics & FD001           & FD002            & FD003             & FD004             \\ \toprule
\multirow{2}{*}{CNN-LSTM}      & RMSE    & 14.78           & 22.11            & 14.65             & 26.69             \\
                               & Score   & 405.30          & 10438.61         & 393.92            & 11123.60          \\ \midrule
\multirow{2}{*}{\begin{tabular}{@{}c@{}}CNN-LSTM\\+MGSC\end{tabular}} & RMSE    & \textbf{12.63}  & \textbf{18.85}   & \textbf{12.78}    & \textbf{25.37}    \\
                               & Score   & \textbf{301.12} & \textbf{1879.65} & \textbf{313.08}   & \textbf{6927.23}  \\ \toprule
\multirow{2}{*}{CNN}           & RMSE    & 14.49           & 21.57            & 14.18             & 25.81             \\
                               & Score   & 396.08          & 3224.94          & 382.21            & \textbf{9882.99}           \\ \midrule
\multirow{2}{*}{CNN+MGSC}      & RMSE    & \textbf{13.65}  & \textbf{20.38}   & \textbf{13.05}  & \textbf{24.85}    \\
                               & Score   & \textbf{280.19} & \textbf{2988.29} & \textbf{286.01} & 14410.83 \\ \toprule
\multirow{2}{*}{LSTM}          & RMSE    & 13.22           & 16.03           & 11.46           & 19.4877            \\
                               & Score   & 292.05          & 1091.71          & 244.77           & 4274.96           \\ \midrule
\multirow{2}{*}{LSTM+MGSC}     & RMSE    & \textbf{12.54}   & \textbf{15.94}    & \textbf{11.33}    & \textbf{18.92}     \\
                               & Score   & \textbf{283.49}  & \textbf{1074.34}  & \textbf{230.08}   & \textbf{1970.88}  \\                               
                               \bottomrule
\end{tabular}
\end{table}

Fig. \ref{fig:{RUL}} shows the RUL prediction results for the four sub-datasets using the CNN-LSTM+MGSC model. The comparison reveals a high similarity between the predicted RUL and the actual RUL, especially in the FD001 and FD003 datasets. This indicates that the proposed model is able to predict the RUL of each engine better without large bias.
\begin{figure}[!t]
  \centering
  \includegraphics[width=0.88\linewidth]{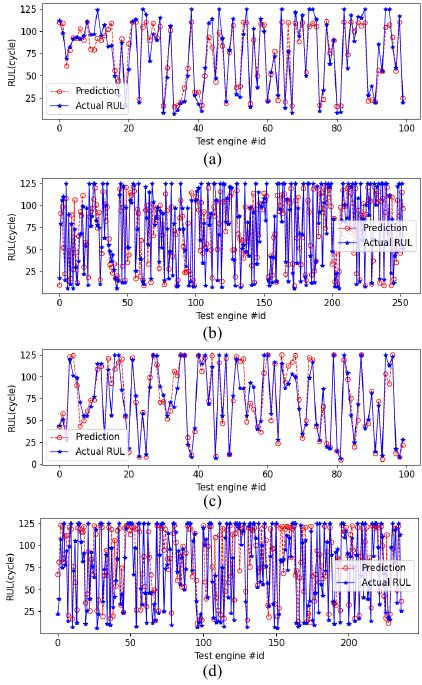}
  \caption{Predicted results of the CNN-LSTM baseline for (a) FD001, (b) FD002,
(c) FD003, and (d) FD004.}
  \label{fig:{RUL}}
\end{figure}

Fig. \ref{fig:{Schematic2}}-Fig. \ref{fig:{Schematic4}} shows the encoder feature space visualization results for the FD002-FD004 dataset with or without the MGSC strategy. The t-SNE results show that the encoder of the used MGSC strategy constitutes an ideal manifold structure in the feature space. For samples of different engines with the same or similar RUL labels, their feature embeddings are close to each other and aligned. This is consistent with our intuition that aligning feature embeddings with the same RUL labels in the feature space before the regression layer will aid in RUL prediction. Furthermore, due to the combined effect of the MSE loss function and the fine-grained contrast loss, there is a clear degradation trend in the feature space, which is good for visualizing and explaining the health state of the aero-engine.

\begin{figure}[!t]
  \centering
  \includegraphics[width=1\linewidth]{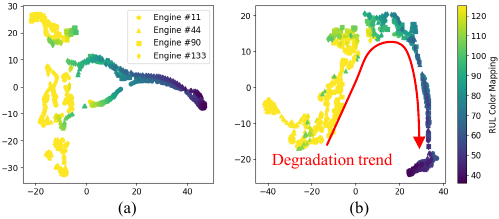}
  \caption{CNN-LSTM Encoder feature space of four test engines from FD002 visualized by t-SNE. (a) Without MGSC. (b) With MGSC.}
  \label{fig:{Schematic2}}
\end{figure}
\begin{figure}[!t]
  \centering
  \includegraphics[width=1\linewidth]{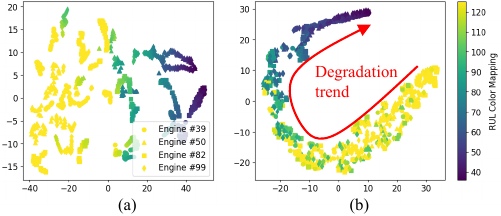}
  \caption{CNN-LSTM Encoder feature space of four test engines from FD003 visualized by t-SNE. (a) Without MGSC. (b) With MGSC.}
  \label{fig:{Schematic3}}
\end{figure}
\begin{figure}[!t]
  \centering
  \includegraphics[width=1\linewidth]{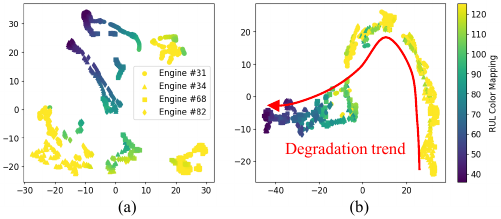}
  \caption{CNN-LSTM Encoder feature space of four test engines from FD004 visualized by t-SNE. (a) Without MGSC. (b) With MGSC.}
  \label{fig:{Schematic4}}
\end{figure}

\section{Conclusion}
In this paper, the research is carried out from the perspective of feature space regularization, and an MGSC strategy is proposed for the alignment of sample embeddings under the same RUL labels, which produces an ideal manifold in the feature space. Inputting this manifold into the regression layer greatly improves the prediction accuracy of RUL.

In MGSC, coarse-grained contrast with HS as class labels, and fine-grained contrast with RUL labels within each HS class as class labels are established, which achieve the interaction between the high-level overall features and low-level local features. Moreover, the issues of large minibatch size and imbalance of positive and negative samples in supervised contrastive learning are solved by different classification scales. The application of the MGSC strategy to the RUL prediction task of an aero-engine is realized by the proposed multi-phase training framework and network structure.

In future work, we will continue to explore multi-granularity contrast learning methods as well as contrast methods that are more applicable to time series. Meanwhile, the MGSC framework will be used for more advanced RUL predictive model to validate the effects.






\bibliographystyle{IEEEtran}
\bibliography{IEEEabrv,mybibfile}

\begin{thebibliography}{1}
\providecommand{\url}[1]{#1}
\csname url@samestyle\endcsname
\providecommand{\newblock}{\relax}
\providecommand{\bibinfo}[2]{#2}
\providecommand{\BIBentrySTDinterwordspacing}{\spaceskip=0pt\relax}
\providecommand{\BIBentryALTinterwordstretchfactor}{4}
\providecommand{\BIBentryALTinterwordspacing}{\spaceskip=\fontdimen2\font plus
\BIBentryALTinterwordstretchfactor\fontdimen3\font minus \fontdimen4\font\relax}
\providecommand{\BIBforeignlanguage}[2]{{%
\expandafter\ifx\csname l@#1\endcsname\relax
\typeout{** WARNING: IEEEtran.bst: No hyphenation pattern has been}%
\typeout{** loaded for the language `#1'. Using the pattern for}%
\typeout{** the default language instead.}%
\else
\language=\csname l@#1\endcsname
\fi
#2}}
\providecommand{\BIBdecl}{\relax}
\BIBdecl

\bibitem{8709986}
X.~Si, T.~Li, and Q.~Zhang, ``A general stochastic degradation modeling approach for prognostics of degrading systems with surviving and uncertain measurements,'' \emph{IEEE Transactions on Reliability}, vol.~68, no.~3, pp. 1080--1100, 2019.

\bibitem{9801529}
Z.~Kong, X.~Jin, Z.~Xu, and B.~Zhang, ``Spatio-temporal fusion attention: A novel approach for remaining useful life prediction based on graph neural network,'' \emph{IEEE Transactions on Instrumentation and Measurement}, vol.~71, pp. 1--12, 2022.

\bibitem{XUE202095}
Z.~Xue, Y.~Zhang, C.~Cheng, and G.~Ma, ``Remaining useful life prediction of lithium-ion batteries with adaptive unscented kalman filter and optimized support vector regression,'' \emph{Neurocomputing}, vol. 376, pp. 95--102, 2020.

\bibitem{9849459}
R.~Jin, M.~Wu, K.~Wu, K.~Gao, Z.~Chen, and X.~Li, ``Position encoding based convolutional neural networks for machine remaining useful life prediction,'' \emph{IEEE/CAA Journal of Automatica Sinica}, vol.~9, no.~8, pp. 1427--1439, 2022.

\bibitem{8967059}
K.~Park, Y.~Choi, W.~J. Choi, H.-Y. Ryu, and H.~Kim, ``Lstm-based battery remaining useful life prediction with multi-channel charging profiles,'' \emph{IEEE Access}, vol.~8, pp. 20\,786--20\,798, 2020.

\bibitem{oord2018representation}
A.~v.~d. Oord, Y.~Li, and O.~Vinyals, ``Representation learning with contrastive predictive coding,'' \emph{arXiv preprint arXiv:1807.03748}, 2018.

\bibitem{weinberger2009distance}
K.~Q. Weinberger and L.~K. Saul, ``Distance metric learning for large margin nearest neighbor classification.'' \emph{Journal of machine learning research}, vol.~10, no.~2, 2009.

\bibitem{chen2020simple}
T.~Chen, S.~Kornblith, M.~Norouzi, and G.~Hinton, ``A simple framework for contrastive learning of visual representations,'' in \emph{International conference on machine learning}.\hskip 1em plus 0.5em minus 0.4em\relax PMLR, 2020, pp. 1597--1607.

\bibitem{khosla2020supervised}
P.~Khosla, P.~Teterwak, C.~Wang, A.~Sarna, Y.~Tian, P.~Isola, A.~Maschinot, C.~Liu, and D.~Krishnan, ``Supervised contrastive learning,'' \emph{Advances in neural information processing systems}, vol.~33, pp. 18\,661--18\,673, 2020.

\end{thebibliography}

\vspace{12pt}

\end{document}